\documentclass[nohyperref]{article}

\usepackage{microtype}
\usepackage{graphicx}
\usepackage{subfigure}
\usepackage{booktabs} 

\usepackage{hyperref}

\usepackage[accepted]{style_files/icml2022}

\usepackage{amsmath}
\usepackage{amssymb}
\usepackage{mathtools}
\usepackage{amsthm}

\usepackage[capitalize,noabbrev]{cleveref}

\theoremstyle{plain}

\theoremstyle{definition}

\theoremstyle{remark}

\usepackage[textsize=tiny]{todonotes}

\usepackage{amsmath,amsfonts,bm}

\def\eqref#1{equation~\ref{#1}}

\def\1{\bm{1}}

\DeclareMathAlphabet{\mathsfit}{\encodingdefault}{\sfdefault}{m}{sl}
\SetMathAlphabet{\mathsfit}{bold}{\encodingdefault}{\sfdefault}{bx}{n}

\newcommand{\E}{\mathbb{E}} 

\DeclareMathOperator*{\argmax}{arg\,max}

\makeatletter
\newcommand{\subalign}[1]{%
  \vcenter{%
    \Let@ \restore@math@cr \default@tag
    \baselineskip\fontdimen10 \scriptfont\tw@
    \advance\baselineskip\fontdimen12 \scriptfont\tw@
    \lineskip\thr@@\fontdimen8 \scriptfont\thr@@
    \lineskiplimit\lineskip
    \ialign{\hfil$\m@th\scriptstyle##$&$\m@th\scriptstyle{}##$\hfil\crcr
      #1\crcr
    }%
  }%
}
\makeatother

\graphicspath{figures/}
\usepackage{xcolor}
\usepackage[export]{adjustbox}
\usepackage{placeins}
\usepackage{footmisc}
\usepackage{url}
\usepackage{float}
\allowdisplaybreaks
\usepackage{todonotes}

\usepackage{hyperref}
\newcommand{\internship}{$\text{}^{\ddagger}$Work conducted while interning at Ubisoft La Forge.}

\icmltitlerunning{Direct Behavior Specification via Constrained Reinforcement Learning}

\begin{document}

\twocolumn[
\icmltitle{Direct Behavior Specification\\via Constrained Reinforcement Learning}



\icmlsetsymbol{intern}{$\ddagger$}

\begin{icmlauthorlist}
\icmlauthor{Julien Roy}{intern,mila,poly}
\icmlauthor{Roger Girgis}{intern,mila,poly}
\icmlauthor{Joshua Romoff}{ubi}
\icmlauthor{Pierre-Luc Bacon}{mila,udem,face}
\icmlauthor{Christopher Pal}{mila,poly,udem,sn,can}
\end{icmlauthorlist}

\icmlaffiliation{mila}{Institut d'intelligence aritficielle du Québec (Mila).}
\icmlaffiliation{poly}{École Polytechnique de Montréal.}
\icmlaffiliation{ubi}{Ubisoft La Forge.}
\icmlaffiliation{udem}{Université de Montréal.}
\icmlaffiliation{sn}{ServiceNow.}
\icmlaffiliation{face}{Facebook CIFAR AI Chair.}
\icmlaffiliation{can}{Canada CIFAR AI Chair.}

\icmlcorrespondingauthor{Julien Roy}{julien.roy@mila.quebec}

\icmlkeywords{Machine Learning, ICML}

\vskip 0.3in
]



\printAffiliationsAndNotice{\internship}  

\begin{abstract}
The standard formulation of Reinforcement Learning lacks a practical way of specifying what are admissible and forbidden behaviors. Most often, practitioners go about the task of behavior specification by manually engineering the reward function, a counter-intuitive process that requires several iterations and is prone to reward hacking by the agent. In this work, we argue that constrained RL, which has almost exclusively been used for safe RL, also has the potential to significantly reduce the amount of work spent for reward specification in applied RL projects. To this end, we propose to specify behavioral preferences in the CMDP framework and to use Lagrangian methods to automatically weigh each of these behavioral constraints. Specifically, we investigate how CMDPs can be adapted to solve goal-based tasks while adhering to several constraints simultaneously. We evaluate this framework on a set of continuous control tasks relevant to the application of Reinforcement Learning for NPC design in video games.
\end{abstract}


\section{Introduction }

Reinforcement Learning (RL) has shown rapid progress and lead to many successful applications over the past few years \cite{mnih2013playing,silver2017mastering,andrychowicz2020learning}. The RL framework is predicated on the simple idea that all tasks could be defined as a single scalar function to maximise, an idea generally referred to as the reward hypothesis \cite{sutton2018reinforcement,silver2021reward,abel2021expressivity}. This idea has proven very useful to develop the theory and concentrate research on a single theoretical framework. However, it can be significantly limiting when translating a real-life problem into an RL problem, since the question of where the reward function comes from is completely ignored~\cite{singh2009rewards}. In practice, human-designed reward functions often lead to unforeseen behaviors and represent a serious obstacle to the reliable application of RL in the industry~\cite{amodei2016concrete}.

Concretely, for an engineer working on applying RL methods to an industrial problem, the task of reward specification implies to: (1) characterise the desired behavior that the system should exhibit, (2) write in a computer program a reward function for which the optimal policy corresponds to that desired behavior, (3) train an RL agent on that task using one of the methods available in the literature and (4) evaluate whether the agent exhibits the expected behavior. Multiple design iterations of that reward function are generally required, each time accompanied by costly trainings of the policy \cite{hadfield2017inverse,dulac2019challenges}. This inefficient design loop is exacerbated by the fact that current Deep RL algorithms cannot be guaranteed to find the optimal policy~\cite{sutton2018reinforcement}, meaning that the reward function could be correctly specified but still fail to lead to the desired behavior. The design problem thus becomes ``What reward function would lead SAC~\cite{haarnoja2018soft} or PPO~\cite{schulman2017proximal} to give me a policy that I find satisfactory?", a difficult puzzle that every RL practitioner has had to deal with.

\begin{figure*}[ht]
    \centering
    \begin{minipage}{.5\textwidth}
        \centering
        \includegraphics[width=.98\textwidth,height=4.3cm]{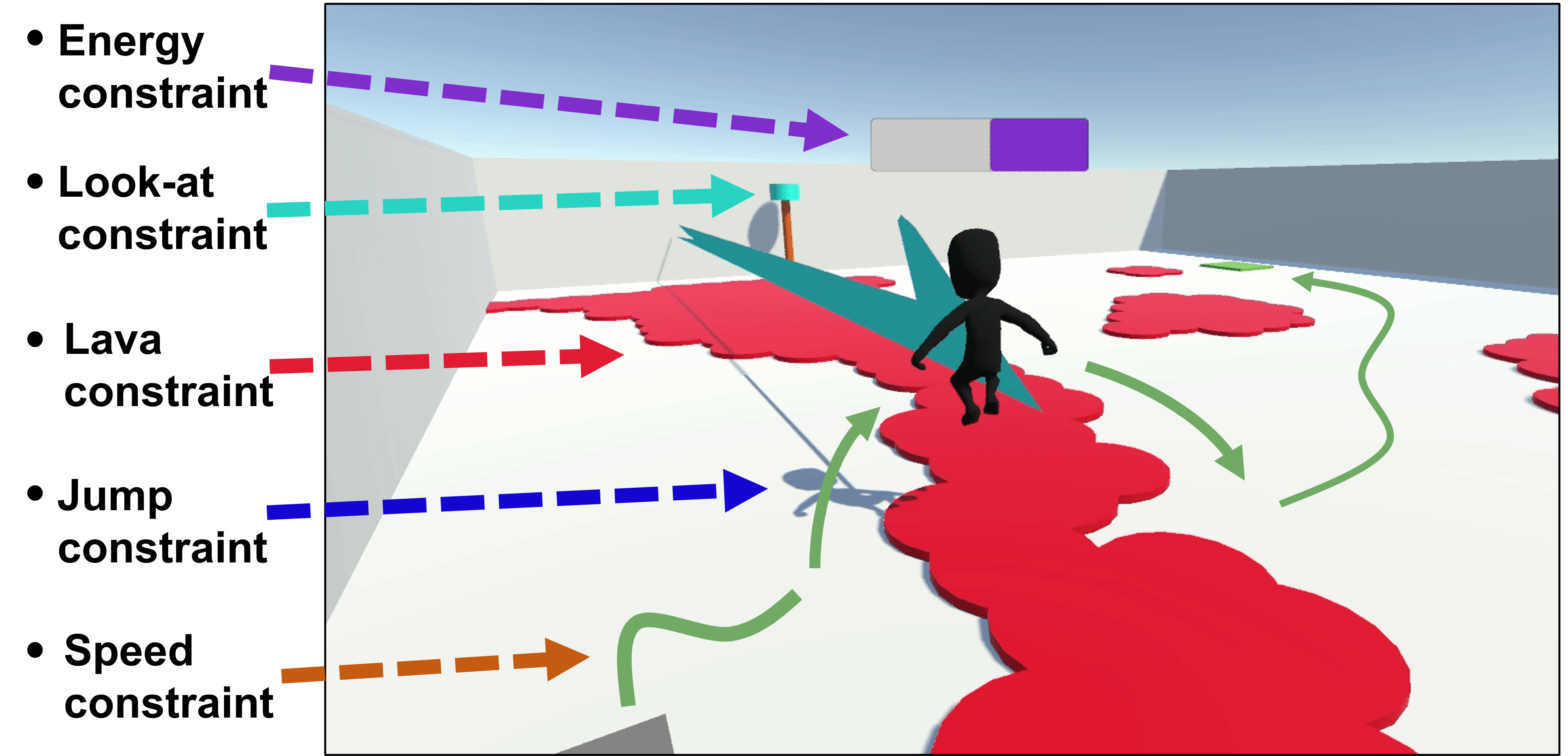}
    \end{minipage}%
    \begin{minipage}{.5\textwidth}
        \centering
        \includegraphics[width=.98\textwidth,height=4.3cm,frame]{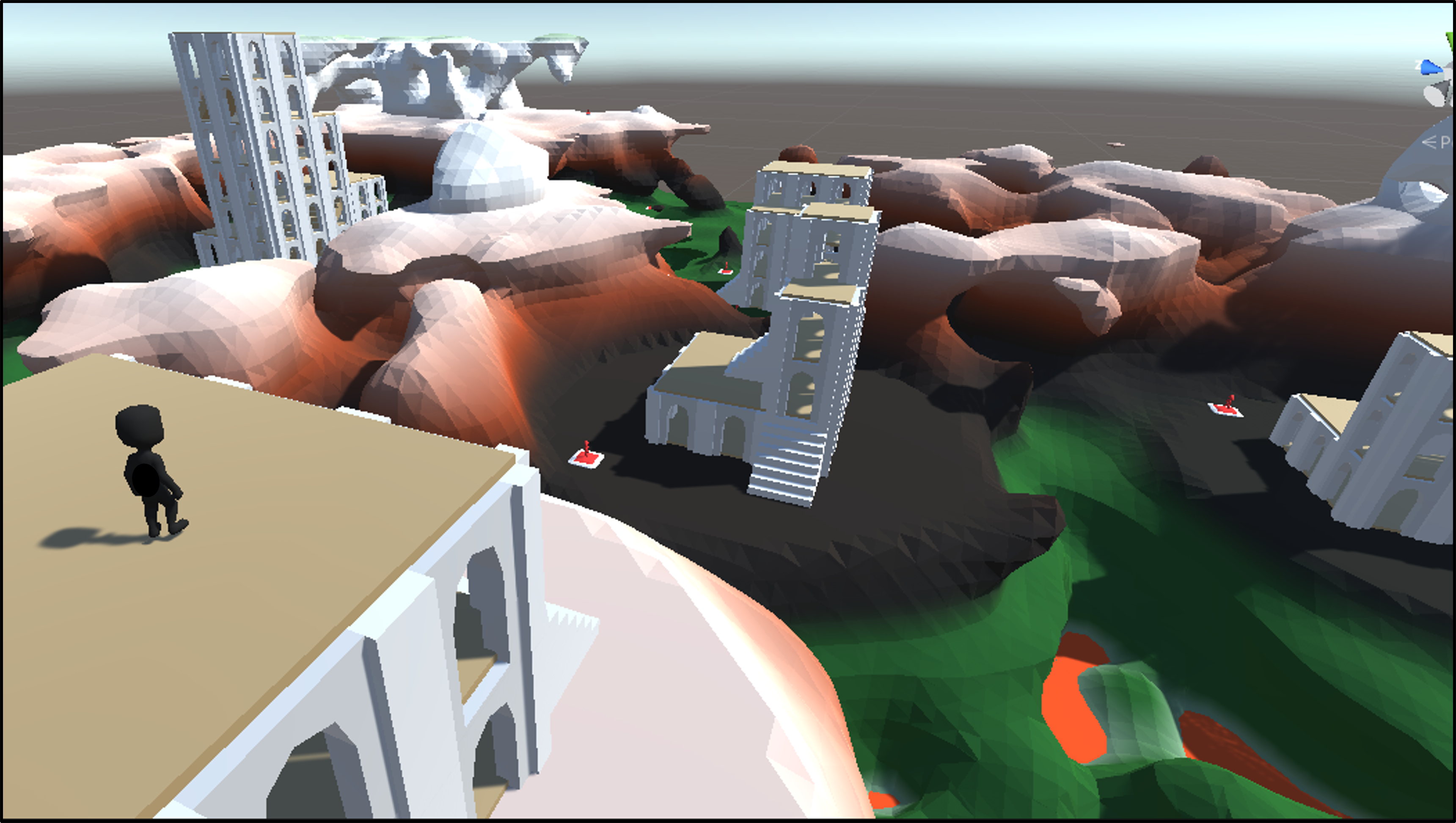}
    \end{minipage}
    \caption{Depictions of our setup to evaluate direct behavior specification using constrained RL; Arena environment (left); OpenWorld environment (right). For videos see: \href{https://sites.google.com/view/behaviorspecificationviacrl/home}{https://sites.google.com/view/behaviorspecificationviacrl/home}.}
    \label{fig:envs}
\end{figure*}

Most published work on Reinforcement Learning focuses on point (3) i.e. improving the reliability and efficiency with which these algorithms can yield a near-optimal policy for a \textit{given} reward function. This line of work is crucial to allow RL to tackle difficult problems. However, as agents become more and more capable of solving the tasks we present them with, our ability to (2) correctly specify these reward functions will only become more critical~\cite{dewey2014reinforcement}.

Constrained Markov Decision Processes~\cite{altman1999constrained} offer an alternative framework for sequential decision making. The agent still seeks to maximise a single reward function, but must do so while respecting a set of constraints defined by additional cost functions. While it is generally recognised that this formulation has the potential to allow for an easier task definition from the end user~\cite{ray2019benchmarking}, most work on CMDPs focuses on the safety aspect of this framework i.e. that the constraint-satisfying behavior be maintained throughout the entire exploration process \cite{achiam2017constrained,zhang2020first,turchetta2020safe,marchesini2022exploring}. In this paper we specifically focus on the benefits of CMDPs relating to behavior specification. We make the following contributions: (1) we show experimentally that reward engineering poorly scales with the complexity of the target behavior, (2) we propose a solution where a designer can directly specify the desired frequency of occurrence of some events, (3) we develop a novel algorithmic approach capable of jointly satisfying many more constraints and (4) we evaluate this framework on a set of constrained tasks illustrative of the development cycle required for deploying RL in video games.

\begin{figure*}[t]
    \centering
    \begin{tabular}{lc}
        \textbf{a)} & 
        \includegraphics[width=0.66\textwidth]{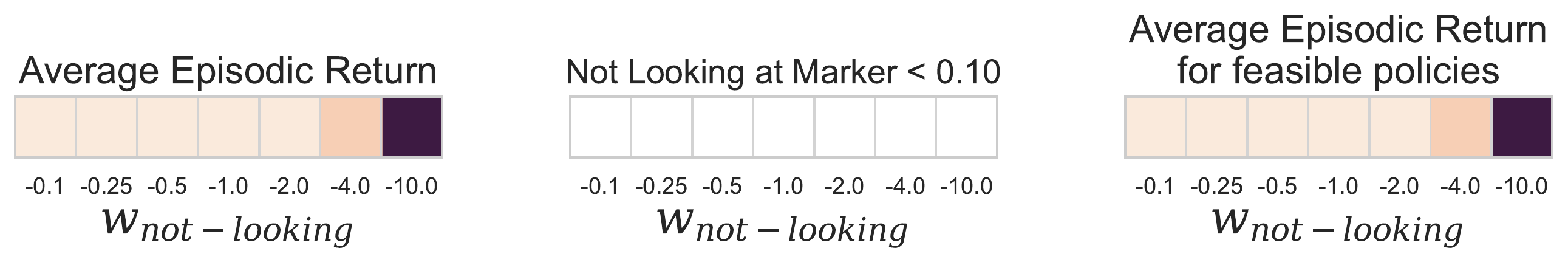}\\
        \textbf{b)} & 
        \includegraphics[width=0.9\textwidth]{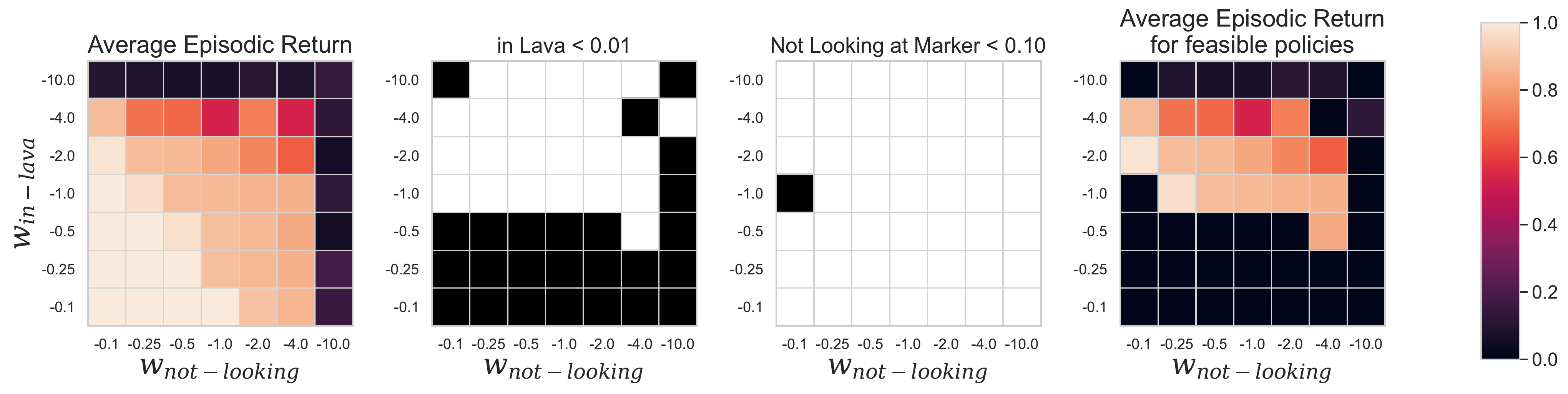}
    \end{tabular}
    \caption{Enforcing behavioral constraints using reward engineering. Each grid represents a different metric. Within each grid, each square represents the final performance (according to that metric) of an agent trained for 3M steps using the reward function in Equation~\ref{eq:reward_engineering} parameterised as given by the grid coordinates. Performance is obtained by evaluating the agent on 1000 episodes. The leftmost column indicates the episodic return of the trained policies, the middle columns indicates whether or not the agent respects the behavioral constraint(s) and the rightmost column indicates the average return for these feasible policies only. \textbf{a)} The ``looking-at marker" behavior does not affect too much the main task and, consequently, all chosen weights allow to satisfy the constraint (looking at marker 90\% of the time) and many of them also lead to good performance on the main navigation task ($-0.1 \geq w \geq -2$). \textbf{b)} When also enforcing the ``Not in Lava" behavior, which is much more in the way of the main task, most of the resulting policies do not respect the constraint or perform poorly on the navigation task, highlighting the difficulty of choosing the correct penalty weights ahead of time. On 49 experiments, only two yielded good performing feasible policies: $(-0.10, -2.0)$ and $(-0.25, -1.0)$. On the largest search with 3 behavioral constraints, none of the 343 experiments found a good performing feasible policy (see Figure~\ref{fig:reward_engineering_3constraints} in Appendix~\ref{sec:additional_experiments}).}
    \label{fig:reward_engineering}
\end{figure*}

\section{The problem with reward engineering}
\label{sec:problem_with_reward_engineering}

In this section, we motivate the impracticality of using reward engineering to shape behavior. 
We consider a navigation task in which the agent has to reach a goal location while being subject to additional behavioral constraints. These constraints are (1) looking at a visible marker $90\%$ of the time, (2) avoiding forbidden terrain $99\%$ of the time and (3) avoiding to run out of energy also $99\%$ of the time. The environment is depicted in Figure~\ref{fig:envs} (left) and the details are presented in Appendix~\ref{sec:arena_env_details}. The reward function for this task is of the form:
\begin{equation}
\label{eq:reward_engineering}
R'(s,a) = R(s,a) - \mathbf{1}*w_{\text{not-looking}} - \mathbf{1} * w_{\text{in-lava}} - \mathbf{1} * w_{\text{no-energy}}
\end{equation}
where $R(s,a)$ gives a small shaping reward for progressing towards the goal and a terminal reward for reaching the goal, and the $\mathbf{1}$'s are indicator functions which are only active if their corresponding behavior is exhibited.

The main challenge for an RL practitioner is to determine the correct values of the weights $w_{\text{not-looking}}$, $w_{\text{in-lava}}$ and $w_{\text{no-energy}}$ such that the agent maximises its performance on the main task while respecting the behavioral requirements, a problem often referred to as reward engineering. Setting these weights too low results in an agent that ignores these requirements while setting them too high distracts the agent from completing the main task. In general, knowing how to scale these components relatively to one another is not intuitive and is often performed by trial and error across the space of reward coefficients $w_k$. To illustrate where the desired solutions can be found for this particular problem, we perform 3 grid searches on 7 different values for each of these weights, ranging from 0.1 to 10 times the scale of the main reward function, for the cases of 1, 2 and 3 behavioral constraints. The searches thus respectively must go through 7, 49 and 343 training runs. Figure~\ref{fig:reward_engineering} (and Figure~\ref{fig:reward_engineering_3constraints} in Appendix~\ref{sec:additional_experiments}) show the results of these experiments. We can see that a smaller and smaller proportion of these trials lead to successful policies as the number of behavioral constraints grows. For an engineer searching to find the right trade-off, they find themselves cornered between two undesirable solutions: an ad-hoc manual approach guided by intuition or to run a computationally demanding grid-search. While expert knowledge or other search strategies can partially alleviate this burden, the approach of reward engineering clearly does not scale as the control problem grows in complexity.

It is important to note that whether or not it is the case that all tasks can in principle be defined as a single scalar function to maximise i.e. the reward hypothesis~\cite{sutton2018reinforcement}, this notion should not be seen as a restrictive design principle when translating a real-life problem into an RL problem. That is because it does not guarantee that this reward function admits a simple form. Rich and multi-faceted behaviors may only be specifiable through a complex reward function \citep{abel2021expressivity} beyond the reach of human intuition. In the next sections we present a practical framework in which CMDPs can be used to provide a more intuitive and human-centric interface for behavioral specification.


\section{Background }
\label{sec:background}

\paragraph{Markov Decision Processes (MDPs)}~\cite{sutton2018reinforcement} are formally defined through the following four components: $(\mathcal{S}, \mathcal{A}, P, R)$. At timestep $t$, an agent finds itself in state $s_t \in \mathcal{S}$ and picks an action $a_t \in \mathcal{A}(s_t)$. The transition probability function $P$ encodes the conditional probability  $P(s_{t+1}|s_t, a_t)$ of transitioning to the next state $s_{t+1}$. Upon entering the next state, an immediate reward is generated through a reward function $R: \mathcal{S} \times \mathcal{A} \to \mathbb{R}$. In this paper, we restrict our attention to stationary randomized policies of the form $\pi(a|s)$ -- which are sufficient for optimality in both MDPs and CMDPs \cite{altman1999constrained}. The interaction of a policy within an MDP gives rise to trajectories $(s_0, a_0, r_0, \hdots, s_T, a_T, R_T)$ over which can be computed the sum of rewards which we call the \textit{return}. Under the Markov assumption, the probability distribution over trajectories is of the form:
\begin{equation}
    p_\pi(\tau) := P_0(s_0)\prod_{t=0}^T P(s_{t+1}|s_t, a_t) \pi(a_t|s_t)
\end{equation}
where $P_0$ is some initial state distribution. Furthermore, any such policy induces a marginal distribution over state-action pairs referred to as the \textit{visitation distribution} or state-action \textit{occupation measure}:
\begin{equation}
    x_{\pi}(s,a) := \frac{1}{Z(\gamma, T)} \sum_{t=0}^T \gamma^t p_{\pi, t}(S_t=s, A_t=a) \enspace ,
\end{equation}
where $Z(\gamma, T) = \sum_{t=0}^T \gamma^t$ is a normalising constant.

In this paper, it is useful to extend the notion of return to any function $f:\mathcal{S}\times \mathcal{A}\rightarrow \mathbb{R}$ over states and actions other than the reward function of the MDP itself. The expected discounted sum of $f$ then becomes:
\begin{equation}
    J_f(\pi) := \E_{\tau \sim p_{\pi}}\left[ \sum_{t=0}^T \gamma^t f(s_t,a_t) \right]
\end{equation}
where $\gamma \in [0,1]$ is a discount factor. While this idea is the basis for much of the work on General Value Functions (GVFs) \cite{white2015} for predictive state representation \citep{sutton2011}, our focus here is on problem of behavior specification and not that of prediction. 

Finally, in the MDP setting, a policy is said to be optimal under the expected discounted return criterion if $\pi^* = \argmax_{\pi \in \Pi} J_R(\pi)$, where $\Pi$ is the set of possible policies. 

\paragraph{Constrained MDPs (CMDPs)}~\cite{altman1999constrained} 
is a framework that extends the notion of optimality in MDPs to a scenario where multiple cost constraints need to be satisfied in addition to the main objective. We write $C_k: \mathcal{S}\times\mathcal{A}\rightarrow \mathbb{R}$ to denote such a cost function whose expectation must remain bounded below a specified threshold $d_k \in \mathbb{R}$. 
The set of feasible policies is then:
\begin{equation}
\Pi_C = \{ \pi  \in \Pi: J_{C_k}(\pi) \leq d_k, \, k=1,\dots,K \}.
\end{equation}
Optimal policies in the CMDP framework are those of maximal expected return among the set of feasible policies:
\begin{equation}
\pi^* = \argmax_{\pi \in \Pi} \, J_R(\pi), \, \text{ s.t. } \, \,  J_{C_k}(\pi) \leq d_k \text{ , } \, \, k=1,\dots,K
\end{equation}
While it is sufficient to consider the space of stationary deterministic policies in searching for optimal policies in the MDP setting, this is no longer true in general with CMDPs \citep{altman1999constrained} and we must consider the larger space of stationary randomized policies. 

\paragraph{Lagrangian methods for CMPDs.}
Several recent works have found that the class of Lagrangian methods for solving CMDPs is capable of finding good feasible solutions at convergence ~\cite{achiam2017constrained,ray2019benchmarking,stooke2020responsive,zhang2020first}. The basis for this line of work stems from the saddle-point characterisation of the optimal solutions in nonlinear programs with inequality constraints \citep{uzawa, polyak, korpelevich1976}. Intuitively, these methods combine the main objective $J_R$ and the constraints into a single function $\mathcal{L}$ called the Lagrangian. The relative weight of the constraints are determined by additional variables $\lambda_k$ called the Lagrange multipliers. Applied in our context, this idea leads to the following min-max formulation:
\begin{align}
\label{eq:cmdp_lagrangian}
&\max_\pi \, \, \min_{\lambda\geq 0} \, \mathcal{L}(\pi, \lambda) \notag \\ &\mathcal{L}(\pi, \lambda) = J_R(\pi) - \sum_{k=1}^K \lambda_k (J_{C_k}(\pi) - d_k)
\end{align}
where we denoted $\lambda:=\{\lambda_k\}_{k=1}^K$ for conciseness. \citet{uzawa} proposed to find a solution to this problem iteratively by taking gradient ascent steps of the Lagrangian $\mathcal{L}$ in the variable $\pi$ and descent ones in $\lambda$. This is also the same gradient ascent-descent 
\cite{gda} procedure underpinning many learning algorithms for Generative Adversarial Networks \citep{goodfellow2014generative}.

The maximization of the Lagrangian over the policy variables can be carried out by applying any existing unconstrained policy optimization methods to the new reward function $L: \mathcal{S} 
\times \mathcal{A} \to \mathbb{R}$ where:
\begin{equation}
L(s,a) = R(s,a) - \sum_{k=1}^K \lambda_k C_k(s,a).
\end{equation}
For the gradient w.r.t. the Lagrange multipliers $\lambda$, the term depending on $\pi$ cancels out and we are left with $\nabla_{\lambda_k} \mathcal{L}(\pi, \lambda) = -(J_{C_k}(\pi) - d_k)$. The update is followed by a projection onto $\lambda_k \geq 0$ using the max-clipping operator. If the constraint is violated $(J_{C_k}(\pi) > d_k)$, taking a step in the opposite direction of the gradient will increase the corresponding multiplier $\lambda_k$, thus increasing the relative importance of this constraint in $J_L(\pi)$. Inversely, if the constraint is respected $(J_{C_k}(\pi) < d_k)$, the update will decrease $\lambda_k$, allowing the optimisation process to focus on the other constraints and the main reward function $R$. 


\section{Proposed Framework}

In Reinforcement Learning, the reward function is often assumed to be provided apriori. For example, in most RL benchmarking environments this is indeed the case and researchers can focus on improving current algorithms at finding better policies, faster and more reliably. In industrial applications however, several desiderata are often required for the agent's behavior, and balancing these components into a single reward function is highly non-trivial. In the next sections, we describe a framework in which CMDPs can be used for efficient behavior specification.

\subsection{Indicator cost functions}

The difficulty of specifying the desired behavior of an agent using a single reward function stems from the need to tune the relative scale of each reward component. Moreover, finding the most appropriate ratio becomes more challenging as the number of reward components increases (see Section~\ref{sec:problem_with_reward_engineering}). While the prioritisation and saturation characteristics of CMDPs help factoring the behavioral specification problem \cite{ray2019benchmarking}, there remains important design challenges. First, the CMDP framework allows for arbitrary forms of cost functions, again potentially leading to unforeseen behaviors. Second, specifying the appropriate thresholds $d_k$ can be difficult to do solely based on intuition. For example, in the mujoco experiments performed by \citet{zhang2020first}, the authors had to run an unconstrained version of PPO \citep{schulman2017proximal} to first estimate the typical range of values for the cost infringements and then run their constrained solver over the appropriately chosen thresholds.

We show here that this separate phase of threshold estimation can be avoided completely if we consider a subclass of CMDPs that allows for a more intuitive connection between the chosen cost functions $C_k$ and their expected returns $J_{C_k}$. More specifically, we restrict our attention to CMDPs where the cost functions are defined as indicators of the form:
\begin{equation}
\label{eq:indicator_cost_functions}
    C_k(s,a) = I(\text{behavior $k$ is met in $(s,a)$})
\end{equation}
which simply expresses whether an agent showcases some particular behavior $k$ when selecting action $a$ in state $s$. An interesting property of this design choice is that, by rewriting the expected discounted sum of these indicator cost functions as an expectation over the visitation distribution of the agent, we can interpret this quantity as a re-scaled probability that the agent exhibits behavior $k$ at any given time during its interactions with the environment:
\begin{align}
    &J_{C_k}(\pi) 
    = \E_{\tau \sim p_{\pi}}\left[ \sum_{t=0}^T \gamma^t C_k(s_t,a_t) \right] \\
    &= Z(\gamma, T) \mathbb{E}_{(s,a)\sim x_{\pi}(s,a)}[C_k(s,a)] \\
    &= Z(\gamma, T) \mathbb{E}_{(s,a)\sim x_{\pi}(s,a)}[I(\text{behavior $k$ met in $(s,a)$})] \\
    &= Z(\gamma, T) Pr\big(\text{behavior $k$ met in $(s,a)$}\big) , \, \, (s,a)\sim x_{\pi}
\end{align}
Dividing each side of $J_{C_k}(\pi) \leq d_k$ by $Z(\gamma, T)$, we are left with $\tilde{d}_k$, a normalized constraint threshold for the constraint $k$ which represents the desired rate of encountering the behavior designated by the indicator cost function $C_k$. In practice, we simply compute the average cost function across the batch to give equal weighting to all state-action pairs regardless of their position $t$ in the trajectory:
\begin{equation}
    \tilde{J}_{C_k}(\pi) := \frac{1}{N}\sum_{i=1}^N C_k(s_i, a_i)
\end{equation}
where $i$ is the sample index from the batch. We also train the corresponding critic $Q^{(k)}$ using a discount factor $\gamma_k<1$ for numerical stability.

While the class of cost functions defined in Equation~\ref{eq:indicator_cost_functions} still allows for modelling a large variety of behavioral preferences, it has the benefit of informing the user on the range of appropriate thresholds -- a probability $\tilde{d}_k\in[0,1]$ -- and the semantics is clear regarding its effect on the agent's behavior (assuming that the constraint is binding and that a feasible policy is found). This effectively allows for minimal to no tuning behavior specification (or ``zero-shot" behavior specification). 

Finally, indicator cost functions also have the practical advantage of allowing to capture both desired and undesired behaviors without affecting the termination tendencies of the agent. Indeed, when using an arbitrary cost function, it could be tempting to simply flip its sign to enforce the opposite behavior. However, as noted in previous work~\cite{kostrikov2018discriminator}, the choice of whether to enforce behaviors through bonuses or penalties should instead be thought about with the termination conditions in mind. A positive bonus could cause the agent to delay termination in order to accumulate more bonuses while negative penalties could shape the agent behavior such that it seeks to trigger the termination of the episode as soon as possible. Indicator cost functions are thus very handy in that they offer a straightforward way to enforce the opposite behavior by simply inverting the indicator function $Not\big(I(s,a)\big) = 1 - I(s,a)$ without affecting the sign of the constraint (penalties v.s. bonuses).

\subsection{Multiplier normalisation}
\label{sec:mult_norm}

When the constraint $k$ is violated, the multiplier $\lambda_k$ associated with that constraint increases to put more emphasis on that aspect of the overall behavior. While it is essential for the multipliers to be able to grow sufficiently compared to the main objective, a constraint that enforces a behavior which is long to discover can end up reaching very large multiplier values. It then leads to very large policy updates and destabilizes the learning dynamics.

To maintain the ability of one constraint to dominate the policy updates when necessary while keeping the scale of the updates bounded, we propose to normalize the multipliers. This can be readily implemented by using a softmax layer:
\begin{equation}
    \lambda_k = \frac{\exp(z_k)}{\exp(a_0) + \sum_{k'=1}^K \exp(z_{k'})} \, , \quad k=1,\dots,K
\end{equation}
where $z_k$ are the base parameters for each one of the multipliers and $a_0$ is a dummy variable used to obtain a normalized weight $\lambda_0 := 1 - \sum_{k=1}^K \lambda_k$ for the main objective $J_R(\pi)$. The corresponding min-max problem becomes:
\begin{align}
\label{eq:cmdp_lagrangian_normalised_multipliers}
&\max_\pi \, \, \min_{z_{1:K}\geq 0} \, \mathcal{L}(\pi, \lambda) \notag \\
&\mathcal{L}(\pi, \lambda) = \lambda_0 J_R(\pi) - \sum_{k=1}^K \lambda_k (J_{C_k}(\pi) - d_k)
\end{align}
\subsection{Bootstrap Constraint}
In the presence of many constraints, one difficulty that emerges with the above multiplier normalisation is that the coefficient of the Lagrangian function that weighs the main objective is constrained to be  $\lambda_0 = 1 - \sum_{k=1}^K \lambda_k$, which leaves very little to no traction to improve on the main task while the process is looking for a feasible policy. Furthermore, as more constraints are added, the optimisation path becomes discontinuous between regions of feasible policies, preventing learning progress on the main task objective.

A possible solution is to grant the main objective the same powers as the behavioral constraints that we are trying to enforce. This can be done by defining an additional function $S_{K+1}(s,a)$ which captures some measure of success on the main task. Indeed, many RL tasks are defined in terms of such sparse, clearly defined success conditions, and then often only augmented with a dense reward function to guide the agent toward these conditions~\cite{ng1999policy}. A so-called \textit{success constraint} of the form $J_{S_{K+1}}(\pi) \geq \tilde{d}_{K+1}$ can thus be implemented using an indicator cost function as presented above and added to the existing constraint set $\{J_{C_k}(\pi) \leq \tilde{d}_{k}\}_{k=1}^K$. While the use of a success constraint alone can be expected to aid learning of the main task, it is only a sparse signal and could be very difficult to discover if the main task is itself challenging. Since the success function $S_{K+1}$ is meant to be highly correlated with the reward function $R$, by going a step further and using the success constraint multiplier $\lambda_{K+1}$ in place of the reward multiplier $\lambda_0$, we can take full advantage of the density of the main reward function when enforcing that constraint. However, to maintain a true maximisation objective over the main reward function, we still need to keep using $\lambda_0$  when other constraints are satisfied, so that the most progress can be made on $J_R(\pi)$. We thus take the largest of these two coefficients for weighing the main objective $\tilde{\lambda}_0 := \max\big(\lambda_0, \lambda_{K+1}\big)$ and replace $\lambda_0$ with $\tilde{\lambda}_0$ in Equation~\ref{eq:cmdp_lagrangian_normalised_multipliers}. Here we say that constraint $K$~$+$~$1$ is used as a \textit{bootstrap constraint}.

Our method of encoding a success criterion in the constraint set can be seen as a way of relaxing the behavioral constraints during the optimisation process without affecting the convergence requirements. For exemple, in previous work, \citet{calian2020balancing} tune the learning rate of the Lagrange multipliers to automatically turn some constraints into soft-constraints when the agent is not able to satisfy them after a given period of time. Instead, the bootstrap constraint allows to start making some progress on the main task without turning our hard constraints into soft constraints.


\section{Related Work}
\label{sec:related_work}

\paragraph{Constrained Reinforcement Learning.} CMDPs~\cite{altman1999constrained} have been the focus of several previous work in Reinforcement Learning. Lagrangian methods~\cite{borkar2005actor,tessler2018reward,stooke2020responsive} combine the constraints and the main objective into a single function and seek to find a saddle point corresponding to feasible solutions to the maximisation problem. Projection-based methods~\cite{achiam2017constrained,chow2019lyapunov,yang2020projection,zhang2020first} instead use a projection step to try to map the policy back into a feasible region after the reward maximisation step. While most of these works focus on the single-constraint case~\cite{zhang2020first,dalal2018safe,calian2020balancing,stooke2020responsive} and seek to minimize the total regret over the cost functions throughout training~\cite{ray2019benchmarking}, we focus on the potential of CMDPs for precise and intuitive behavior specification and work on satisfying many constraints simultaneously.

\paragraph{Reward Specification.} Imitation Learning~\cite{zheng2021imitation} is largely motivated by the difficulty of designing reward functions and instead seeks to use expert data to define the task. Other approaches introduce a human in the loop to either guide the agent towards the desired behavior~\cite{christiano2017deep} or to prevent it from making catastrophic errors while exploring the environment~\cite{saunders2017trial}. While our approach of using CMDPs for behavior specification also seeks to make better use of human knowledge, we focus on the idea of providing this knowledge by simply specifying thresholds and indicator functions rather than requiring expert demonstrations or constant human feedback. Another line of work studies whether natural language can be used as a more convenient interface to specify the agent's desired behavior~\cite{goyal2019using,macglashan2015grounding}. While this idea presents interesting perspectives, natural language is inherently ambiguous and prone to reward hacking by the agent. Moreover such approaches generally come with the added complexity of having to learn a language-to-reward model. Finally, others seek to solve reward mis-specification through Inverse Reward Design~\cite{hadfield2017inverse,mindermann2018active,ratner2018simplifying} which treats the provided reward function as a single observation of the true intent of the designer and seeks to learn a probabilistic model that explains it. While this approach is interesting for adapting to environmental changes, we focus on behavior specification in fixed-distribution environments.

\paragraph{RL in video games.} Video games have been used as a benchmark for Deep RL for several years \cite{shao2019survey,berner2019dota,vinyals2019grandmaster}. However, examples of RL being used in a video game production are limited due to a variety of factors which include the difficulty of shaping behavior, interpretability, and compute limitations at run-time \cite{jacob2020s,alonso2020deep}. Still, there has been a recent push in the video game industry to build NPCs (Non Player Characters) using RL, for applications including navigation \cite{alonso2020deep,devlin2021navigation}, automated testing \cite{bergdahl2020augmenting,gordillo2021improving}, play-style modeling \cite{de2021configurable} and content generation \cite{gisslen2021adversarial}.


\begin{figure*}
    \centering
    \includegraphics[width=0.9\textwidth]{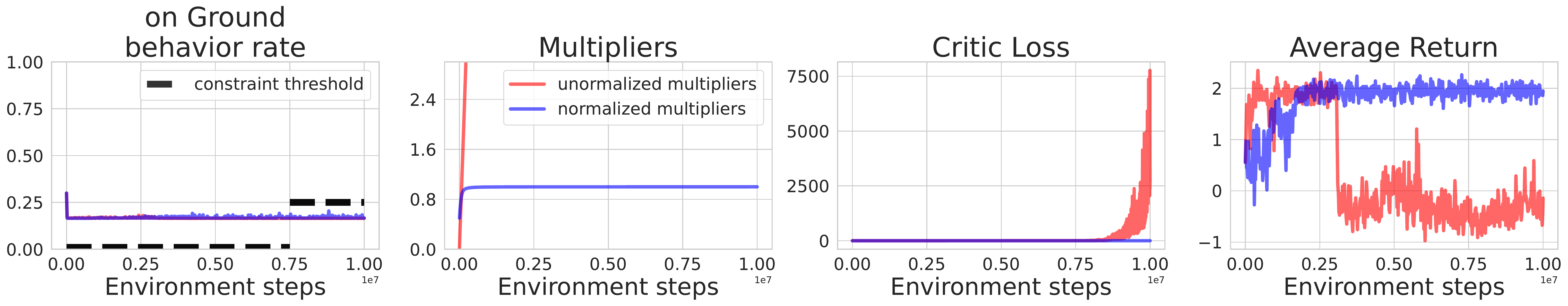}
    \caption{The multiplier normalisation keeps the learning dynamics stable when discovering a constraint-satisfying behavior takes a large amount of time. To simulate such a case, an impossible constraint is set for 7.5M steps and then replaced by a feasible one for the last 2.5M steps. The method using unormalized multipliers (red) keeps taking larger and larger steps in policy space leading to the divergence of its learning dynamics and complete collapse of its performance.}
    \label{fig:multiplier_normalisation_experiment}
\end{figure*}
\begin{figure*}
    \centering
    \includegraphics[width=0.9\textwidth]{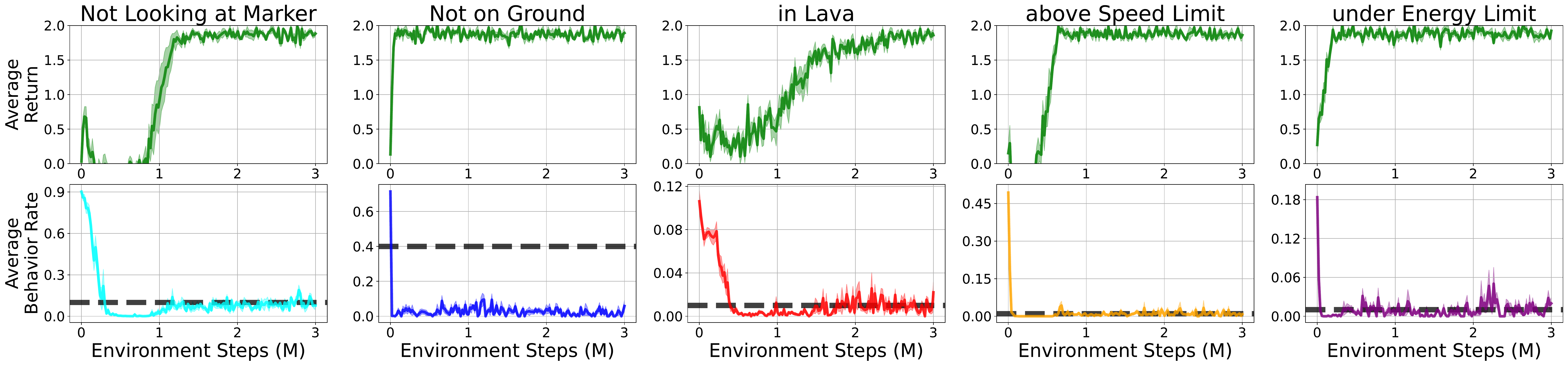}
    \caption{Each \textit{column} presents the results for an experiment in which the agent is trained for 3M steps with a \textit{single constraint} enforced on its behavior. Training is halted after every $20,000$ environment steps and the agent is evaluated for 10 episodes. All curves show the average over 5 seeds and envelopes show the standard error around that mean. The top row shows the average return, the bottom row shows the average behavior rate on which the constraint is enforced. The black doted lines mark the constraint thresholds.}
    \label{fig:single_constraint_experiments}
\end{figure*}
\begin{figure*}
    \centering
    \includegraphics[width=\textwidth]{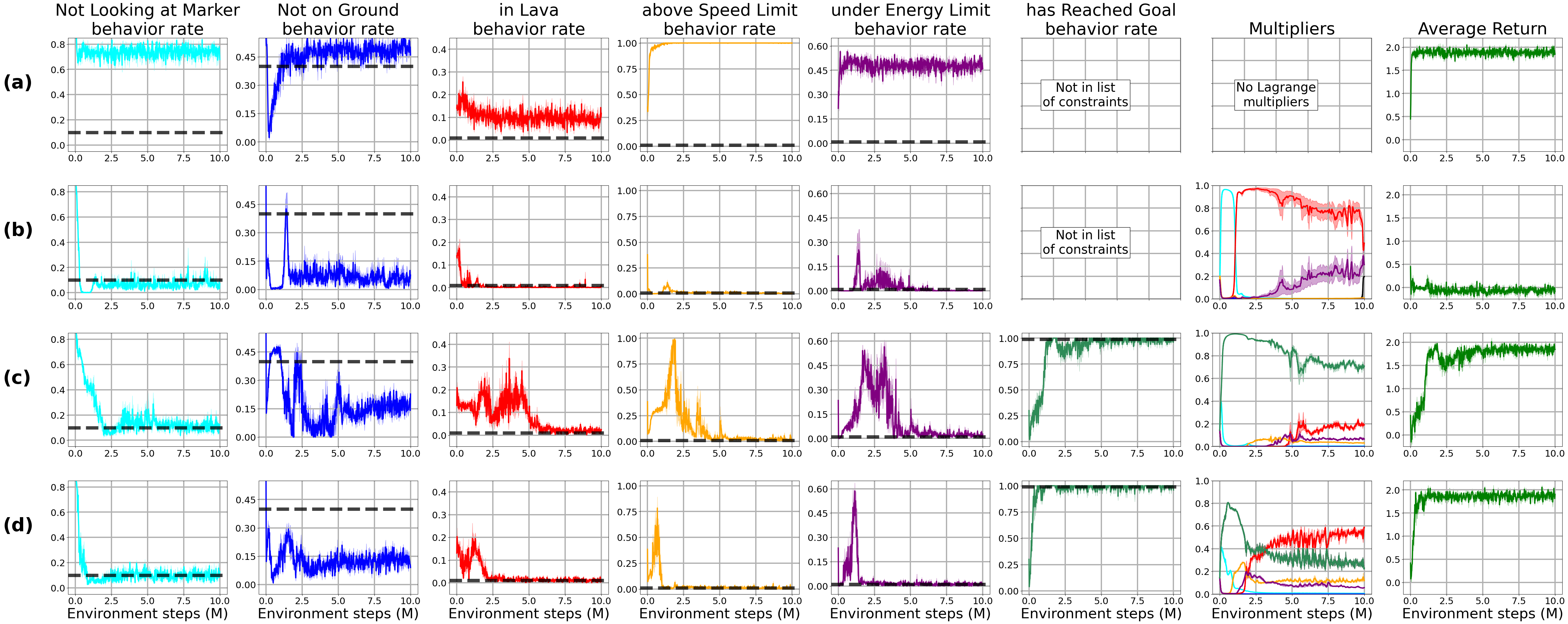}
    \caption{Each \textit{row} presents the results of an experiment in which an agent is trained for 10M steps. Training is halted after every $20,000$ environment steps and the agent is evaluated for 10 episodes.  All curves show the average over 5 seeds and envelopes show the standard error around that mean. \textbf{(a)} Unconstrained SAC agent; none of the behavioral preferences are enforced and consequently improvement on performance is very fast but none of the constraints are satisfied. \textbf{(b)} SAC-Lagrangian with the 5 behavioral constraints enforced. While each constraint was successfully dealt with when imposed one by one (see Figure~\ref{fig:single_constraint_experiments}), maximising the main objective when subject to all the constraints \textit{simultaneously} proves to be much harder. The agent does not find a policy that improves on the main task while keeping the constraints in check. \textbf{(c)} By using an additional success constraint (that the agent should reach its goal in 99\% of episodes), the agent can cut through infeasible policy space to start improving on the main task and optimise the remaining constraints later on. \textbf{(d)} By using the success constraint as a bootstrap constraint (bound to the main reward function) improvement on the main task is much faster as the agent benefits from the dense reward function to improve on the goal-reaching task.}
    \label{fig:many_constraints_experiments}
\end{figure*}

\begin{figure*}[h!t]
    \centering
    \includegraphics[width=\textwidth]{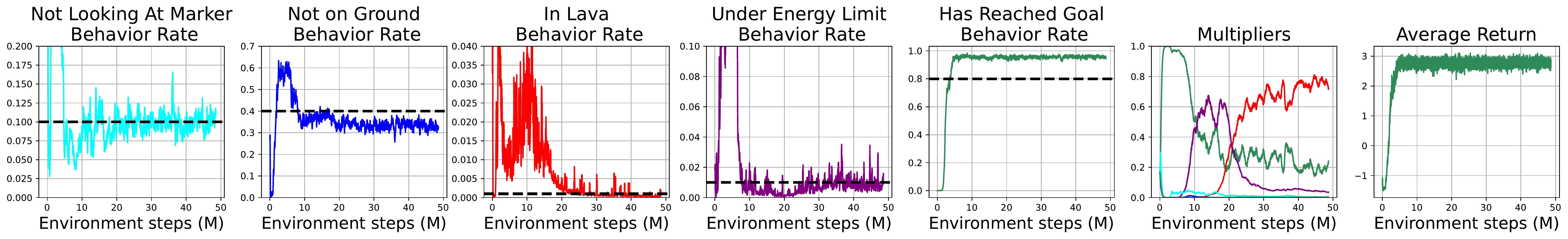}
    \caption{A SAC-Lagrangian agent trained to solve the navigation problem in the OpenWorld environment while respecting four constraints and imposing the bootstrap constraint. Results suggest that our SAC-Lagrangian method using indicator cost functions, normalised multipliers and bootstrap constraint scales well to larger and more complex environments.}
    \label{fig:open_world_exp}
\end{figure*}

\section{Experiments}

To evaluate the proposed framework, we train SAC agents~\cite{haarnoja2018soft} to solve navigation tasks with up to 5 constraints imposed on their behavior. Many of these constraints interact with the main task and with one another which significantly restricts the space of admissible policies. We conduct most of our experiments in the Arena environment (see Figure~\ref{fig:envs}, left)\footnote{The algorithm is presented in Appendix~\ref{sec:algorithm}. The code for the Arena environment experiments is available at:\newline\scriptsize{\url{https://github.com/ubisoft/DirectBehaviorSpecification}}} where we seek to verify the capacity of the proposed framework to allow for easy specification of the desired behavior and the ability of the algorithm to deal with a large number of constraints simultaneously. We also perform an experiment in the OpenWorld environment (see Figure~\ref{fig:envs}, right), a much larger and richer map generated using the GameRLand map generator~\cite{beeching2021graph}, where we seek to verify the scalability of that approach and whether it fits the needs of agent behavior specification for the video game industry. See Appendices~\ref{sec:arena_env_details}~and~\ref{sec:openworld_env_details} for a detailed description of both experimental setups.

\subsection{Experiment in the Arena environment}

\paragraph{Multiplier Normalization}

Our first set of experiments showcases the effect of normalizing the Lagrange multipliers. For illustrative purposes, we designed a simple scenario where one of the constraints is not satisfied for a long period of time. Specifically, the agent is attempting to satisfy an impossible constraint of never touching the ground. Figure~\ref{fig:multiplier_normalisation_experiment} (in red) shows that the multiplier on the unsatisfied constraint endlessly increases in magnitude, eventually harming the entire learning system; the loss on the critic diverges and the performance collapses. When using our normalization technique, Figure~\ref{fig:multiplier_normalisation_experiment} (in blue) shows that the multiplier and critic losses remain bounded, avoiding such instabilities.

\paragraph{Single Constraint satisfaction}

We use our framework to encode the different behavioral preferences into indicator functions and specify their respective thresholds. Figure \ref{fig:single_constraint_experiments} shows that our SAC-Lagrangian with multiplier normalisation can solve the task while respecting the behavioral requirements when imposed with constraints individually. We note that the different constraints do not affect the main task to the same extent; while some still allow to quickly solve the navigation task, like the behavioral requirement to avoid jumping, others make the navigation task significantly more difficult to solve, like the requirement to avoid certain types of terrain (lava).

\paragraph{Multiple Constraints Satisfaction}

In Figure \ref{fig:many_constraints_experiments} we see that when imposed with all of the constraints simultaneously, the agent learns a feasible policy but fails at solving the main task entirely. The agent effectively settles on a trivial behavior in which it only focuses on satisfying the constraints, but from which it is very hard to move away without breaking the constraints. By introducing a success constraint, the agent at convergence is able to satisfy all of the constraints as well as succeeding in the navigation task. This additional incentive to traverse infeasible regions of the policy space allows to find feasible but better performing solutions. Our best results are obtained when using the success constraint as a bootstrap constraint, effectively lending $\lambda_{K+1}$ to the main reward while the agent is still looking for a feasible policy.

\subsection{Experiment in the OpenWorld environment}

In the OpenWorld environment, we seek to verify that the proposed solution scales well to more challenging and realistic tasks. Contrarily to the Arena environment, the OpenWorld contains uneven terrain, buildings, and interactable objects like jump-pads, which brings this evaluation setting much closer to an actual RL application in the video game industry. For this experiment, we trained a SAC-Lagrangian agent to solve the navigation problem with four constraints on its behavior: \textit{On-Ground}, \textit{Not-In-Lava}, \textit{Looking-At-Marker} and \textit{Above-Energy-Limit}. The SAC component uses the same hyper-parameters as in \citet{alonso2020deep}. The results are shown in Figure~\ref{fig:open_world_exp}. While training the agent in this larger and more complex environment now requires up to 50M environment steps, the agent still succeeds at completing the task and respecting the constraints, favourably supporting the scalability of the proposed framework for direct behavior specification. 


\section{Discussion}

Our work showed that CMDPs offer compelling properties when it comes to task specification in RL. More specifically, we developed an approach where the agent's desired behavior is defined by the frequency of occurrence for given indicator events, which we view as constraints in a CMDP formulation. We showed through experiments that this methodology is preferable over the reward engineering alternative where we have to do an extensive hyper-parameter search over possible reward functions. We evaluated this framework on the many constraints case in two different environments. Our experiments showed that simultaneously satisfying a large number of constraints is difficult and can systematically prevent the agent from improving on the main task. We addressed this problem by normalizing the constraint multipliers, which resulted in improved stability during training and proposed to bootstrap the learning on the main objective to avoid getting trapped by the composing constraint set. 
This bootstrap constraint becomes a way for practitioners to incorporate prior knowledge about the task and desired result -- if the threshold is strenuous, a high success is prioritized -- if the threshold is lax, it will simply be used to exit the initialisation point and the other constraints will quickly takeover.
Our overall method is easy to implement over existing policy gradient code bases and can scale across domains  easily. 

We hope that these insights can contribute to a wider use of Constrained RL methods in industrial application projects, and that such adoption can be mutually beneficial to the industrial and research RL communities.


\section*{Acknowledgments}

We wish to thank Philippe Marcotte, Maxim Peter, Rémi Labory, Pierre Le Pelletier De Woillemont, Julien Varnier, Pierre Falticska, Gabriel Robert, Vincent Martineau, Olivier Pomarez, Tristan Deleu and Paul Barde as well as the entire research team at Ubisoft La Forge for providing technical support and insightful comments on this work. We also acknowledge funding in support of this work from Fonds de Recherche Nature et Technologies (FRQNT), Mitacs Accelerate Program, Institut de valorisation des données (IVADO) and Ubisoft La Forge.



\bibliography{sources.bib}
\bibliographystyle{style_files/icml2022}

\newpage
\appendix
\onecolumn

\section{Algorithm}
\label{sec:algorithm}

Our implementation of the SAC-Lagrangian algorithm is presented below. The exact values of each hyper-parameter for all of our experiments are listed in Tables~\ref{table:hyperparams_arena}~and~\ref{table:hyperparams_openWorld}. One notable difference between an unconstrained Soft-Actor Critic~\cite{haarnoja2018soft} and our constrained version is that SAC is typically updated after every environment step to maximise the sample efficiency of the algorithm. In the constrained case however, since the constraints are optimized on-policy, updating the SAC agent at every environment step would only allow for one-sample estimates of the multiplier's objective. On the other hand, freezing the SAC-agent for as many environment steps as the Lagrange multiplier batch-size $N_\lambda$ makes the overall algorithm significantly less sample efficient. One \textit{could} disregard the ``on-policyness" of the multiplier's objective but in preliminary experiments we found that, unsurprisingly, updating the Lagrange multipliers very frequently while using a large set of samples (many of which were collected using previous versions of the policy) lead to significant overshoot and harms the ability of the multipliers to converge to a stable behavior. There is thus a tradeoff to make between the variance of the multiplier's objective estimate, the degree to which the multipliers are updated on-policy and the sample efficiency of the overall algorithm. In practice we found that the values for $M_\theta$ and $M_\lambda$ presented in Tables~\ref{table:hyperparams_arena}~and~\ref{table:hyperparams_openWorld} represented good compromises between these different characteristics. Another important detail is that we use $K+1$ separate critics to model the discounted expected sum of reward and costs. $Q^{(0)}$ is the critic that models the main objective and $Q^{(k)}, k=1,\dots,K+1$ are the critics that model the constraint components of the Lagrangian. Using separate critics allows to avoid fast changes in the scale of the objective, as seen by the critics, when the multipliers $\lambda_k$ get adjusted; they can solely focus on modeling the agent's changing behavior with respect to their respective function (reward or costs).

\begin{algorithm*}
\small
    \begin{algorithmic}
    \caption{\label{alg:SAC-Lagrangian}SAC-Lagrangian with Bootstrap Constraint}
    \REQUIRE learning rate $\beta$, replay buffer $\mathcal{B}$, entropy coefficient $\alpha$ and minibatch sizes $N_\theta$ and $N_\lambda$
    \REQUIRE Initialise the policy $\pi_\theta$ and value-functions $Q^{(k)}_{\phi}$ randomly, $k=0,\dots,K+1$
    \REQUIRE Initialise the Lagrange multiplier parameters $z_k$
    \REQUIRE Collect enough transitions to fill $\mathcal{B}$ with $max(N_\theta, N_\lambda)$ samples
    \FOR{updates $u=1,...$ (until convergence)}
        \STATE \textcolor{gray}{\textbf{\# Data collection}}
        \STATE Sample from the current policy: $a\sim \pi_{\theta}(\cdot|s)$
        \STATE Query next state, reward and indicators $(s', r, \{e\}_{k=1}^{K+1})$  by interacting with the environment
        \STATE Append transition $(s, a, r, s', \{e\}_{k=1}^K+1)$ to the replay buffer $\mathcal{B}$
        \STATE \textcolor{gray}{\textbf{\# Policy Gradient update}}
        \IF{$u \, \% \, M_{\theta} == 0$}
        \STATE Sample a minibatch of $N_\theta$ transitions \textbf{uniformly} from the replay buffer
        \STATE Sample next actions: $\qquad a_i' \sim \pi_\theta(\cdot|s_i') \quad i=1,...,N_{\theta}$
        \FOR{$k=0,\dots,K+1$}
            \STATE Set the ``rewards" to their corresponding values: $\qquad r^{(0)}_i = r_i\quad$ and $\quad r^{(k)}_i = e_i^{(k)}$
            \STATE Compute the Q-targets: $\qquad y_i^{(k)} = -\alpha \log \pi_\theta(a_i'|s_i') + \min_{j \in \{1,2\}} Q_{\phi_j}^{(k)}(s_i', a_i')$
            \STATE Adam descent on Q-nets with: $\qquad \nabla_{\phi_j}\frac{1}{N_\theta}\sum_{i=1}^{N_\theta}||Q_{\phi_j}^{(k)}(s_i, a_i) - \big(r_i^{(k)} + (1 - done) \gamma y_i^{(k)}\big)||_2$
        \ENDFOR
        \STATE Re-sample the current actions: $\qquad a_i \sim \pi_\theta(\cdot|s_i) \quad i=1,...,N_\theta$
        \STATE Adam ascent on policy with: $$\qquad \nabla_{\theta} \frac{1}{N_\theta}\sum_{i=1}^{N_{\theta}} - \alpha \log \pi_\theta(a_i|s_i) + \max(\lambda_0, \lambda_{K+1})\min_{j}Q_{\phi_j}^{(0)}(s_i, a_i) +\lambda_{K+1} \min_{j}Q_{\phi_j}^{(K+1)}(s_i, a_i) 
        - \sum_{k=1}^{K} \lambda_k \min_{j}Q_{\phi_j}^{(k)}(s_i, a_i)$$
        \ENDIF
        \STATE \textcolor{gray}{\textbf{\# Multipliers update}}
        \IF{$u \, \% \, M_\lambda == 0$}
        \STATE Draw from the replay buffer a minibatch composed of \textbf{the last} $N_{\lambda}$ transitions
        \FOR{$k=0,\dots,K+1$}
        \STATE Compute average costs: $\qquad \tilde{J}_{C_k}(\pi) = \frac{1}{N_{\lambda}}\sum_{i=1}^{N_\lambda}e_i^{(k)}$
        \STATE Adam descent on multipliers with: $\qquad \nabla_{z_k} \lambda_k (\tilde{J}_{C_k}(\pi) - \tilde{d}_k) \, \, $ if $\, \, k = K+1 \, \,$ else $\, \, \nabla_{z_k} \lambda_k (\tilde{d}_k - \tilde{J}_{C_k}(\pi))$
        \ENDFOR
        \ENDIF
    \ENDFOR
  \end{algorithmic}
\end{algorithm*}
\normalsize

\clearpage
\section{Details for experiments in the Arena environment}
\label{sec:arena_env_details}

\subsection{Environment details}

In the Arena Environment, the agent's main goal is to navigate to the green tile (see Figure~\ref{fig:envs}, left). The constraints that we explore in this environment are \{\textit{On-Ground}, \textit{Not-in-Lava}, \textit{Looking-At-Marker}, \textit{Under-Speed-Limit} and \textit{Above-Energy-Limit}\}. It receives as observations its XYZ position, direction and velocity, the relative XZ position of the goal, its distance to the goal, as well as an indicator for whether it is on the ground. For the looking-at constraint, it also receives the XZ vector for the direction it is looking at, its Y-angular velocity, the marker's relative XZ position and distance, the normalised angle between the agent's looking direction and the marker as well as an indicator for whether the marker is within its field of view (a fixed-angle cone in front of the agent). For the energy constraint, the agent receives the normalised value of its energy bar and an indicator for whether it is currently recharging. Finally for the lava constraint, the agent receives an indicator of whether it currently stands in lava as well as an indicator for 25 vertical raycast of its surrounding (0 indicating safe ground and 1 indicating lava). We also add to the agent's observations the per-episode rates of indicator cost functions to the agent observation for each of the constraint as well a normalised representation of the remaining time-steps before reaching the time limit condition, leading to a total dimensionality of 53 for the observation vector. The action space is composed of 5 continuous actions (clamped between -1 and 1) which represent its XZ velocity and Y-angular velocity, a jump action (jump is triggered when the agent outputs a value above 0 for that dimensionality) and a recharge action (also with threshold of 0). The reward function is simply 1 when the agent reaches the goal (causing termination), 0 otherwise, and augmented with a small shaping reward function~\cite{ng1999policy} based on whether the agent got closer or further away from the goal location.

\subsection{Hyper-parameters}

Most of the hyper-parameters are the same as in the original unconstrained Soft Actor-Critic (SAC)~\cite{haarnoja2018soft}. Some additional hyper-parameters emerge from the constraint enforcement aspect of our version of SAC-Lagrangian and are described in the Algorithm section above. We use the Adam optimizer~\cite{kingma2014adam} for all parameter updates (policy, critics and Lagrange multipliers). For all experiments taking place in the Arena Environment, the policy is parameterized as a a two layer neural networks that outputs the parameters of a Gaussian distribution with a diagonal covariance matrix. The hidden layers are composed of 256 units and followed by a $tanh$ activation function. The first hidden layer also uses layer-normalisation before the application of the $tanh$ function. We use $K + 1$ fully independent critic models to estimate the expected discount sum of each of the constraint and of the main reward function. The critic models are also parameterized with two-hidden-layers neural networks with the same size for the hidden layers as the policy but instead followed by $relu$ activation functions. Table \ref{table:hyperparams_arena} shows the hyper-parameters used in our experiments conducted in the Arena environment.

\begin{table*}[!hb]
\small
\centering
\begin{sc}
\caption{Hyper-parameters for experiments in the Arena Environment.}
\begin{tabular}{llr}
\label{table:hyperparams_arena}
\\ \toprule
\textbf{General} &
\hspace{5mm} Discount factor $\gamma$                          &{0.9} \\
&\hspace{5mm} Number of random exploration steps     &{10000} \\\vspace{1.2mm}
&\hspace{5mm} Number of buffer warmup steps          &{2560} \\
\textbf{SAC Agent} &
\hspace{5mm} Learning rate $\beta$                    &{0.0003} \\
&\hspace{5mm} Transitions between updates $M_\theta$              &{200} \\
&\hspace{5mm} Batch size $N_\theta$                    &{256} \\
&\hspace{5mm} Replay buffer size                       &{1,000,000} \\
&\hspace{5mm} Initial entropy coefficient $\alpha$             &{0.02} \\\vspace{1.2mm}
&\hspace{5mm} Target networks soft-update coefficient $\tau$             &{0.005} \\
\textbf{Lagrange Multipliers} &
\hspace{5mm} Learning rate $\beta$                    &{0.03} \\
&\hspace{5mm} Initial multiplier parameters value $z_k$                    &{0.02} \\
&\hspace{5mm} Transitions between updates $M_\lambda$               &{2000} \\\vspace{1.2mm}
&\hspace{5mm} Batch size $N_\lambda$                    &{2000} \\
\textbf{Constraint Thresholds} &
\hspace{5mm} Has reached goal (lower-bound)                       &0.99 \\
&\hspace{5mm} \textbf{NOT} looking at marker                       &0.10 \\
&\hspace{5mm} \textbf{NOT} on ground                               &0.40 \\
&\hspace{5mm} In lava                                              &0.01 \\
&\hspace{5mm} Above speed limit                                    &0.01 \\
&\hspace{5mm} Is under the minimum energy level                    &0.01 \\
\bottomrule
\end{tabular}
\end{sc}
\end{table*}

\clearpage
\section{Details for experiments in the OpenWorld environment}
\label{sec:openworld_env_details}

\subsection{Environment details}


The OpenWorld environment is a large environment (approximately $30,000$ times larger than the agent) that includes multiple multi-storey buildings with staircases, mountains, tunnels, natural bridges and lava.
In addition, the environment includes $50$ jump-pads that propel the agent into the air when it steps on one of them.
The agent is tasked with navigating towards a goal randomly placed in the environment at the beginning of every episode.
The agent controls include translation in the XY frame ($2$ inputs), a jumping action ($1$ input), a rotation action controlling where the agent is looking independent of its direction of travel ($1$ input), and a recharging action which allows the agent to recharge its energy level ($1$ input).
The recharging action immobilizes the agent, i.e., it does not allow the agent to progress towards its goal.
The environment also includes a look-at marker which we would like the agent to look at while it accomplishes its main navigation task.

At every timestep, the agent receives as observations its XYZ position relative to the goal as well as its normalized velocity and acceleration in the environment.
In addition, it receives its relative position to the nearest jump-pad in the environment.
For looking at the marker, as in the Arena environment, the agent receives the marker's relative XZ position and distance, the normalised angle between the agent's looking direction and the marker, as well as an indicator for whether the marker is within its field of view (a fixed-angle cone in front of the agent).
For the energy-limit constraint, the agent obtains the value of its energy level, a boolean describing if it is currently recharging and a Boolean indicating if it was recharging in the previous timestep.
The agent also receives a series of indicators denoting whether it is currently standing in lava, if it is touching the ground, and if the agent is currently below the minimum energy level.
In order for the agent to observe lava and other elements it can collide with in the environment (e.g., buildings, doors, mountains), the agent receives 2 channels of $8\times8$ raycasts around the agent.

\subsection{Hyper-parameters}

The SAC agent in the OpenWorld environment uses the same architecture and similar hyper-parameters as in \cite{alonso2020deep}.
The raycasts and raw state described above are processed using two separate embedding models.
For the raycasts, we employ a CNN with 3 convolutional layers, each with a corresponding ReLU layer.
The raw state is processed using a separate 3-layer MLP with $1024$ hidden units at each layer.
The two representations are concatenated into a single vector representing the current state.
The policy is parameterized by a 3-layer MLP that receives as input the concatenated representation and outputs the parameters of a Gaussian distribution with a diagonal covariance matrix. 
Each hidden layer is composed of $1024$ hidden units and is followed by a ReLU activation function.
The critic models are also parameterized by 3-layer MLP, are composed of $1024$ hidden units and use ReLU activation functions.
Table \ref{table:hyperparams_openWorld} shows some of these hyper-parameters with a focus on the constrained enforcement aspect of our version of SAC-Lagrangian.

\begin{table*}[h!]
\small
\centering
\begin{sc}
\caption{Hyper-parameters for experiments in the OpenWorld Environment.}
\begin{tabular}{llr}
\label{table:hyperparams_openWorld}
\\ \toprule
\textbf{General} &
\hspace{5mm} Discount factor $\gamma$                          &{0.99} \\
&\hspace{5mm} Number of random exploration steps $\beta$     &{200} \\\vspace{1.2mm}
&\hspace{5mm} Number of buffer warmup steps $\beta$          &{2560} \\
\textbf{SAC Agent} &
\hspace{5mm} Learning rate $\beta$                    &{0.0001} \\
&\hspace{5mm} Batch size $N_\theta$                    &{2560} \\
&\hspace{5mm} Replay buffer size                       &{4,000,000} \\
&\hspace{5mm} Initial entropy coefficient $\alpha$             &{0.005} \\\vspace{1.2mm}
&\hspace{5mm} Target networks soft-update coefficient $\tau$             &{0.005} \\ 
\textbf{Lagrange Multipliers} &
\hspace{5mm} Learning rate $\beta$                    &{0.00005} \\
&\hspace{5mm} Initial multiplier parameters value $z_k$                    &{0.02} \\
&\hspace{5mm} Transitions between updates                & every timestep \\\vspace{1.2mm}
&\hspace{5mm} Batch size $N_\lambda$                    &{5000} \\
\textbf{Constraint Thresholds} &
\hspace{5mm} Has reached goal (lower-bound)                 &0.80 \\
&\hspace{5mm} \textbf{NOT} looking at marker                 &0.10 \\
&\hspace{5mm} \textbf{NOT} on ground                         &0.40 \\
&\hspace{5mm} In lava                                        &0.001 \\
&\hspace{5mm} Is under the minimum energy level              &0.01 \\
\bottomrule
\end{tabular}
\end{sc}
\end{table*}


\clearpage
\section{Additional experiments on reward engineering}
\label{sec:additional_experiments}

See Section~\ref{sec:problem_with_reward_engineering} for the description of our experiments motivating against the use of reward engineering for behavior specification. Figure~\ref{fig:reward_engineering_3constraints} below shows the results for the biggest of the 3 grid searches performed to showcase the difficulty of finding a reward function that fits the behavioral requirements when the number of requirements grows.

\begin{figure*}[h!b]
    \centering
    \includegraphics[width=0.92\textwidth]{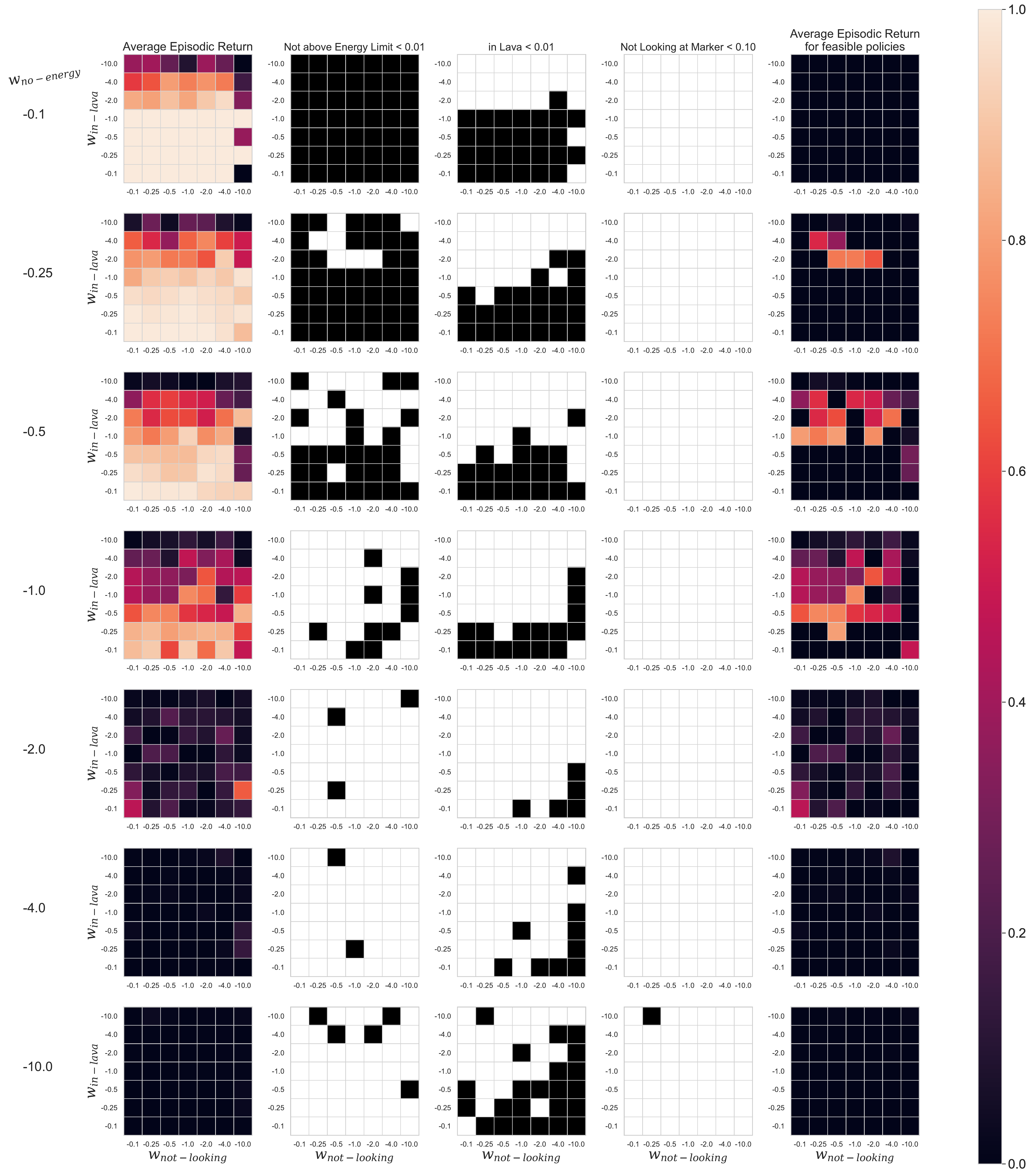}
    \caption{Also see Figure~\ref{fig:reward_engineering}. When enforcing 3 behavioral requirements with reward engineering, an ever larger proportion of the experiments are wasted finding either low-performing policies or policies that do not satisfy the behavioral constraints. In this case, none of the 343 experiments yielded a feasible policy that also solves the task (success rate near 1.0), showcasing that reward engineering scales poorly with the number of constraints due to the curse of dimensionality and to the composing effect of the multiple constraints in narrowing the space of feasible policies.}
    \label{fig:reward_engineering_3constraints}
\end{figure*}


\clearpage
\section{Additional experiments on TD3}
\label{sec:additional_experiments_td3}

\begin{figure*}[h!b]
    \centering
    \includegraphics[trim={3cm 0 0 0},clip,width=\textwidth]{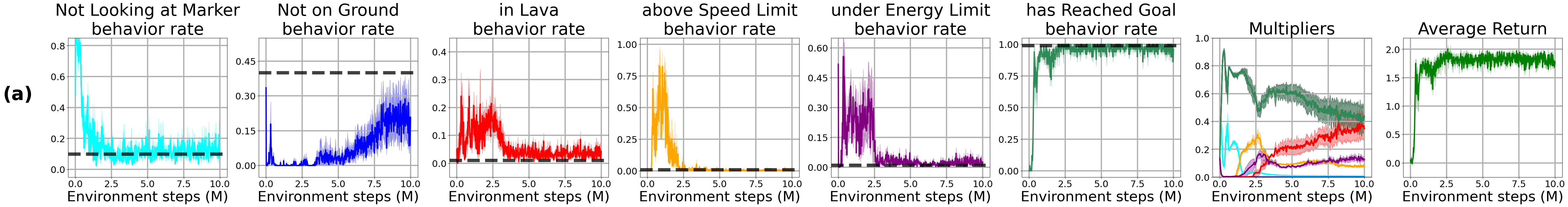}
    \caption{TD3-Lagrangian agent in the Arena environment using normalised multipliers, indicator cost functions and using the success constraint as a bootstrap constraint. Training is halted after every $20,000$ environment steps and the agent is evaluated for 10 episodes. All curves show the average over 5 seeds and envelopes show the standard error around that mean.}
    \label{fig:td3_all_constraints}
\end{figure*}

We validate that our framework can be combined with any policy optimisation algorithm by applying it to the TD3 algorithm~\cite{fujimoto2018addressing}. This leads to a TD3-Lagrangian formulation using our indicator cost functions, normalized multipliers and bootstrap constraint. As for our experiments with SAC (Figure~\ref{fig:many_constraints_experiments}-d), our TD3-Lagrangian agent performs well and all constraints are satisfied. The results are presented in Figure~\ref{fig:td3_all_constraints}.

\end{document}